\documentclass[twocolumn,10pt]{article}
\usepackage[letterpaper,margin=1.8cm,columnsep=0.6cm]{geometry}
\usepackage{amsmath,amssymb}
\usepackage{newtxtext,newtxmath}
\usepackage{microtype}
\usepackage{graphicx}
\usepackage{tikz}
\usetikzlibrary{arrows.meta,calc}
\usepackage{algorithm}
\usepackage{algpseudocode}
\usepackage{caption}
\usepackage[hidelinks]{hyperref}

\makeatletter
\renewcommand\section{\@startsection{section}{1}{\z@}%
  {-2.0ex \@plus -1ex \@minus -.2ex}{1.1ex \@plus.2ex}{\normalfont\large\bfseries}}
\renewcommand\subsection{\@startsection{subsection}{2}{\z@}%
  {-1.6ex \@plus -1ex \@minus -.2ex}{0.8ex \@plus .2ex}{\normalfont\normalsize\bfseries}}
\renewcommand{\paragraph}{\@startsection{paragraph}{4}{\z@}%
  {0.9ex \@plus .4ex \@minus .2ex}{-1em}%
  {\normalfont\normalsize\bfseries}}
\newcommand{\algwrapmargin}{\the\ALG@thistlm}
\makeatother

\definecolor{genblue}{HTML}{5B8FC9}
\definecolor{anchora}{HTML}{E69F00}
\definecolor{anchred}{HTML}{D55E00}
\definecolor{okgreen}{HTML}{2E7D32}

\newcommand{\ifs}{\mathrm{IFS}}
\newcommand{\ivs}{\mathrm{IVS}}
\newcommand{\anch}{\mathcal{A}}

\title{\Large\bfseries Adaptive Identity Anchoring: Closed-Loop Keyframe Placement\\
for Synthetic Paired Supervision in Video Face Swapping}
\author{Logan Robbins$^{*}$}
\date{July 2026}

\begin{document}

\twocolumn[{%
\maketitle
\vspace{-2.5em}
\begin{center}\begin{minipage}{0.9\textwidth}
\footnotesize\textbf{Abstract.}
Video face swapping has no natural paired supervision: no real footage exists of one person's face performing another person's video. The strongest current answer, DreamID-V's SyncID-Pipe, mints pairs by replacing the identity in exactly two frames of a real clip---the first and the last---and regenerating the rest from a pose sequence alone. Pose carries no appearance evidence of the swapped-in identity, so over long clips, occlusions, and extreme pose excursions the synthesized identity has a long unanchored span on which to drift; no published ablation examines anchor count or placement. We propose \emph{Adaptive Identity Anchoring} (AIA): (i)~generalize the synthesizer to arbitrary anchor sets, architecturally natural for diffusion-forcing-style transformers where conditioning on a frame is clamping its tokens to zero noise; (ii)~place anchors by a closed feedback loop that scores every generated frame against the \emph{real} reference identity and inserts an image-face-swapped anchor at the worst-scoring frame until the pair passes a threshold or exhausts a budget; (iii)~reuse the loop's verdict as an automatic data filter. A second pathology, the beauty-filter look of over-smoothed skin, has the same root cause: micro-texture, like identity, is priced by none of the pipeline's objectives. We therefore pair AIA with Reality-Referenced Texture Restoration: matched re-graining from each real frame's non-face regions, band-split transfer of sub-identity micro-texture from the real footage, and a second, spectral acceptance channel refereed by the footage's own spectrum. Identity-anchor density, we argue, is a controllable quality dial, and we specify falsifiable experiments---drift-versus-gap curves, uniform-versus-adaptive placement at matched budgets, student training on AIA-minted data, and texture ablations with a human beauty-filter study---that would validate or refute the proposal.
\end{minipage}\end{center}
\vspace{0.8em}
}]
{\renewcommand{\thefootnote}{\fnsymbol{footnote}}\footnotetext[1]{Independent researcher. \texttt{ljrweb@gmail.com}}}

\section{Introduction}
\label{sec:intro}

Video generation improves on a self-supervised diet: every clip ever recorded is its own training pair, so progress is limited mainly by data collection and compute. Video \emph{transfer} tasks enjoy no such luxury. Video face swapping (VFS) must reproduce identity $B$'s face on a performance that only identity $A$ ever gave, and the ideal ground-truth video---$B$'s face, $A$'s performance, $A$'s scene---does not exist. Supervision must be manufactured, and the fidelity of the manufactured pairs, rather than model capacity, becomes the binding constraint on what a student model can learn. This asymmetry is, we argue, the central reason pure generation has outrun transfer: swap models trained without explicit pairs~\cite{vividface,hifivfs,canonswap} must reconcile identity injection and attribute preservation implicitly, and their identity similarity has lagged that of image face swapping (IFS)~\cite{dreamid}.

DreamID-V~\cite{dreamidv} confronts the missing-pair problem directly. Its data factory, SyncID-Pipe, pre-trains an Identity-Anchored Video Synthesizer (IVS): a first--last-frame-to-video (FLF2V) foundation model with a trainable Adaptive Pose-Attention adapter injected into frozen diffusion-transformer (DiT)~\cite{dit} blocks, conditioned on the first and last frames of a real clip and on its extracted pose sequence. To mint a training pair, a state-of-the-art IFS model~\cite{dreamid} swaps identity $B$ onto the two boundary frames of a real video $V_r$, and the IVS regenerates the interior, yielding a synthetic video $V_g$ of identity $B$ performing $V_r$'s motion. Each mint is used bidirectionally, and---critically---in the direction that swaps $A$ into $V_g$, the regression target is the \emph{real} video $V_r$: real pixels bound how much of the teacher-generator's manifold bias the student inherits.

Our starting observation is that in this factory, appearance identity enters each minted video at exactly \emph{two} temporal sites, fixed a priori at the clip boundaries. Everything in between is constrained only by the pose sequence, which is dense in time but carries no appearance evidence of $B$: DreamID-V's own expression-adaptation step deliberately retargets the landmark sequence through a 3D face reconstruction precisely so that the dense conditioning does not leak source identity~\cite{dreamidv}. Between the anchors, the synthesizer is free to drift back toward its own manifold on the one attribute---identity---for which no dense reference exists. Consistent with this reading, DreamID-V reports that identity similarity remains high in frontal views and mild motion but degrades considerably during profile views and intense actions, and its final Identity-Coherence Reinforcement Learning (IRL) stage exists to re-weight training toward exactly those low-similarity segments~\cite{dreamidv}. Its published ablations cover the quadruplet construction, the curriculum stages, and IRL; none varies anchor count or anchor placement.

We propose \emph{Adaptive Identity Anchoring} (AIA). The one-line thesis: \emph{in a synthetic paired-data factory for video face swapping, identity-anchor density is a controllable quality dial, and anchors should be placed adaptively---by a closed feedback loop scored against the real reference identity---rather than fixed at the clip boundaries.} Concretely, we propose to (a)~train the IVS with randomized anchor sets so that any anchor count and placement is in-distribution at inference; (b)~mint each pair by iteratively scoring the generated video per frame against the real reference photo of $B$ with an identity encoder~\cite{arcface} and inserting an IFS-swapped anchor at the worst-scoring frame, subject to a guard window, an acceptance threshold $\tau$, and an anchor budget $K$; and (c)~treat clips that fail to converge within budget as automatically flagged hard examples or rejects.

The closed-loop principle generalizes beyond identity. The same pipelines are structurally predisposed to a second, older pathology---the beauty-filter look of skin stripped of pores, fine wrinkles, and grain~\cite{deepfacelab}---and the diagnosis is the same: micro-texture, like identity, is an axis on which no factory objective ever consults real pixels (\S\ref{sec:beautyfilter}). We therefore pair AIA with \emph{Reality-Referenced Texture Restoration} (RTR): re-grain and re-texture minted faces from the real footage itself, and referee acceptance through a second, spectral channel (\S\ref{sec:texture}). Our contributions:

\begin{itemize}
\setlength\itemsep{0.15em}
\item \textbf{$N$-anchor synthesis.} A generalization of the two-anchor IVS to arbitrary anchor sets, argued to be architecturally near-free for DiT synthesizers under per-frame noise-level training~\cite{diffusionforcing}, with randomized anchor-set training as the enabling recipe (\S\ref{sec:nanchor}).
\item \textbf{Closed-loop placement.} A formal algorithm that spends a fixed anchor budget exactly where measured identity drift is worst, doubles as an automatic data filter, and emits a per-pair, machine-checkable identity certificate (\S\ref{sec:loop}--\S\ref{sec:integration}).
\item \textbf{Texture honesty.} A five-cause diagnosis of the beauty-filter pathology (\S\ref{sec:beautyfilter}) and Reality-Referenced Texture Restoration---matched re-grain, band-split micro-texture transfer, and a spectral acceptance channel refereed by the footage's own spectrum---in the same mint loop (\S\ref{sec:texture}).
\item \textbf{Analysis and a falsifiable programme.} A constraint-density account of why anchor placement should control drift and why identity and texture are the two axes that need external referees (\S\ref{sec:theory}), and proposed experiments specified concretely enough to execute, each with an ex-ante hypothesis and the interpretation of a negative result (\S\ref{sec:experiments}).
\end{itemize}

\section{Background}
\label{sec:background}

\subsection{SyncID-Pipe and DreamID-V}
\label{sec:syncid}

DreamID-V~\cite{dreamidv} is, to our knowledge, the first DiT-based VFS framework, with released models built on the Wan2.1-1.3B video foundation model~\cite{wan}. Because VFS has no natural pairs, its central contribution is the SyncID-Pipe data factory. An FLF2V foundation model is augmented with a lightweight pose guider and an Adaptive Pose-Attention mechanism: in each DiT block, queries, keys, and values from the frozen backbone are combined with a second attention branch over pose features through trainable key/value projections, with a scalar controlling pose strength and shared rotary position indices for spatiotemporal alignment. Trained with flow matching~\cite{flowmatching} to reconstruct portrait videos from their boundary frames and pose sequence, this becomes the IVS.

Minting proceeds as follows: given a real video $V_r$ of identity $A$ and a reference image of identity $B$, an IFS model~\cite{dreamid} swaps $B$ onto the first and last frames of $V_r$; the two swapped frames plus a retargeted pose sequence (identity coefficients from $B$'s image, expression and pose from each $V_r$ frame, re-projected to landmarks) drive the IVS to synthesize $V_g$. The mint yields a bidirectional ID quadruplet $\{I_r, V_r, I_g, V_g\}$ used in both swap directions; the direction swapping $A$ into $V_g$ regresses onto real pixels $V_r$, and an enhanced-background recomposition pastes generated foregrounds over real backgrounds so that supervision remains real wherever possible. The student is then trained on a synthetic-to-real curriculum, followed by IRL: a full sampling pass is scored per frame by the inverse ArcFace~\cite{arcface} similarity between the generated face and the target identity image, scores are averaged per VAE chunk, and the flow-matching loss is re-weighted by these chunk scores on data selected for high identity variance. DreamID-V reports that IRL raises mean ArcFace similarity on its IDBench-V benchmark from 0.631 to 0.659 while reducing the frame-wise similarity variance from 0.0041 to 0.0029~\cite{dreamidv}---direct published evidence that the residual failure mode of the pipeline is temporally localized identity inconsistency.

\subsection{Anchored synthesis and keyframe interpolation}
\label{sec:anchored}

Conditioning video generation on boundary frames is an established capability: the open Wan2.1 family ships a dedicated FLF2V variant~\cite{wan}, and keyframe interpolation with large image-to-video models has been studied directly~\cite{geninbetween}. Generative Inbetweening~\cite{geninbetween} documents the failure mode any anchored synthesizer must handle: satisfying the anchors as pixel constraints is not the same as producing coherent \emph{motion through} them; naively fusing a forward generation path from one keyframe with a time-reversed path from the other yields back-and-forth motion between the anchors---content that reaches each keyframe without moving coherently through the interval---which that work addresses with dual-directional diffusion sampling that keeps the two paths' motion estimates consistent. Separately, Diffusion Forcing~\cite{diffusionforcing} establishes that sequence diffusion models can be trained with an independent noise level per token, after which ``conditioning on a frame'' is simply clamping that frame's tokens to zero noise---a mechanism indifferent to which frames are clamped. These two lines jointly suggest that anchor \emph{count and position} are exposed, trainable degrees of freedom of the synthesizer, not fixed architectural facts.

\subsection{Identity encoders as referees}
\label{sec:encoders}

ArcFace~\cite{arcface} embeddings with cosine similarity are the de facto identity metric in face swapping, and DreamID-V already employs ArcFace as the reward signal inside IRL and, together with additional encoders such as CurricularFace~\cite{curricularface}, as its evaluation instrument~\cite{dreamidv}. AIA uses the same class of encoder, but earlier in the pipeline: as the referee of a data-minting loop rather than of student training. Pose extraction for the conditioning stream is likewise standard; DreamID-V's released variants use MediaPipe or DWPose~\cite{dwpose}.

\subsection{A second gap: the beauty-filter pathology}
\label{sec:beautyfilter}

Alongside identity drift, we argue the modern factory is structurally predisposed to a second, older pathology: over-smoothed skin---no pores, fine wrinkles, or grain---the beauty-filter look. The observation predates diffusion: DeepFaceLab's authors note that faces from state-of-the-art swapping are ``smoothed and lack minor details (i.e., moles, wrinkles)'' and countered with super-resolution and an optional adversarial critic whose ablations sharpen exactly those details~\cite{deepfacelab}---historically, an adversarial loss was the one term that priced texture in. Modern factories dropped the critic, and the smoothing is now over-determined; we identify five compounding causes.
(a)~\emph{No pixel budget.} The video VAE through which both the IVS and the student operate discards high-frequency micro-texture---learned reconstruction and synthesis models systematically underrepresent high frequencies~\cite{ffl}---and a face occupying a small crop of a 480p--720p frame has little pixel budget for pores at all.
(b)~\emph{Regression to the conditional mean.} Denoising- and flow-matching-style training is mean-squared regression of a target field; wherever sampling under-resolves the residual stochasticity---few-step and distilled samplers above all---outputs slide toward the conditional mean, and the mean of many plausible micro-texture realizations is smooth skin. The IFS teacher that mints the anchors is itself a single-step distilled model~\cite{dreamid}.
(c)~\emph{Temporal objectives reward deletion.} Real facial detail is two superimposed stochastic fields with different motion attachments: sensor grain is static in screen space, while pores and fine wrinkles travel with the face. Rendering both coherently is hard, and a texture-free face scores better on warp-error and smoothness metrics than an almost-right textured one. No temporal-consistency term appears in this factory's losses or rewards; the pressure enters through evaluation---IDBench-V scores motion smoothness~\cite{dreamidv}---and, we argue, the model selection such benchmarks steer: temporal objectives price texture \emph{out} wherever they operate.
(d)~\emph{Nothing prices texture in.} ArcFace-style embeddings are trained for recognition invariance across blur, resolution, and imaging conditions, and are consequently blind to micro-texture; the pipeline's only training signals---latent-space flow-matching regression and the ArcFace-based identity reward---leave texture unpriced, and its evaluation instruments (pose and expression error, motion smoothness) do not measure it either.
(e)~\emph{The flywheel.} Students train partly on teacher output (the quadruplet's synthetic direction), so each generation inherits and, we hypothesize, amplifies its teacher's smoothing.
Anchors alone address none of these; \S\ref{sec:texture} does.

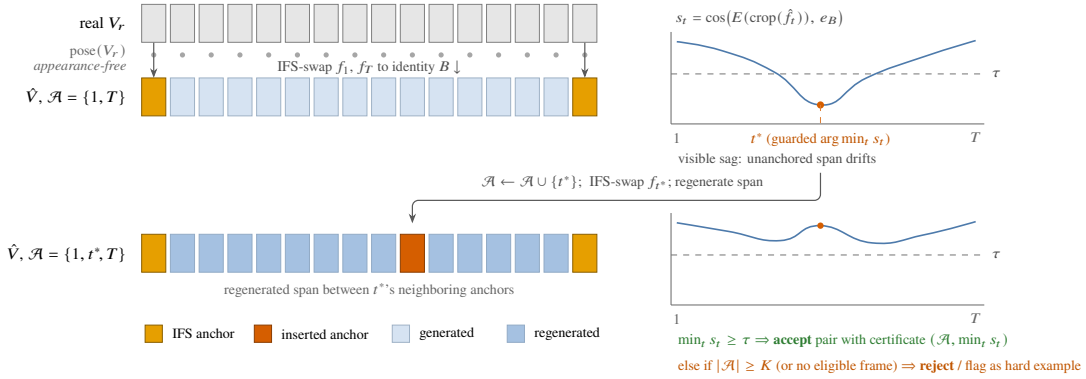
\begin{figure*}[t]
\centering
\resizebox{0.8\textwidth}{!}{%
\begin{tikzpicture}[x=1cm,y=1cm,
  fr/.style={draw=black!50, fill=black!10, minimum width=4.2mm, minimum height=6.6mm, inner sep=0pt},
  gfr/.style={fr, fill=genblue!25, draw=genblue!70!black!50},
  refr/.style={fr, fill=genblue!55, draw=genblue!80!black!60},
  an/.style={fr, fill=anchora, draw=anchora!60!black},
  nan/.style={fr, fill=anchred, draw=anchred!65!black},
  lab/.style={font=\footnotesize},
  slab/.style={font=\scriptsize},
  arr/.style={-{Stealth[length=1.7mm]}, semithick, black!65}]

\foreach \i in {0,...,15} {\node[fr] (r\i) at (1.9+0.5*\i, 5.9) {};}
\foreach \i in {0,...,15} {\fill[black!35] (1.9+0.5*\i, 5.35) circle (0.045);}
\node[lab, anchor=east] at (1.55, 5.9) {real $V_r$};
\node[slab, anchor=east, black!60] at (1.55, 5.42) {pose$(V_r)$};
\node[slab, anchor=east, black!60] at (1.55, 5.14) {\emph{appearance-free}};

\draw[arr] (r0.south) -- ($(1.9,4.62)+(0,0.33)$);
\draw[arr] (r15.south) -- ($(9.4,4.62)+(0,0.33)$);
\node[slab, black!70] at (5.65, 5.13) {$\ifs$-swap $f_1$, $f_T$ to identity $B$ $\downarrow$};

\node[an]  (m0)  at (1.9, 4.62) {};
\foreach \i in {1,...,14} {\node[gfr] (m\i) at (1.9+0.5*\i, 4.62) {};}
\node[an]  (m15) at (9.4, 4.62) {};
\node[lab, anchor=east] at (1.55, 4.62) {$\hat V$, $\anch=\{1,T\}$};

\draw[black!60] (10.9,4.15) -- (16.35,4.15);
\draw[black!60] (10.9,4.15) -- (10.9,5.75);
\node[slab, anchor=south west, black!70] at (10.86, 5.72) {$s_t=\cos\!\big(E(\mathrm{crop}(\hat f_t)),\, e_B\big)$};
\draw[dashed, black!55] (10.9,5.02) -- (16.35,5.02) node[slab, anchor=west, black!70] {$\tau$};
\draw[genblue!85!black, thick] plot [smooth, tension=0.85] coordinates
  {(11.0,5.58) (11.8,5.42) (12.7,5.05) (13.5,4.48) (14.3,4.95) (15.3,5.32) (16.2,5.60)};
\fill[anchred] (13.5,4.48) circle (0.062);
\draw[dashed, anchred!80] (13.5,4.48) -- (13.5,4.15);
\node[slab, anchred!90!black, anchor=north] at (13.5,4.12) {$t^{*}$ (guarded $\arg\min_t s_t$)};
\node[slab, black!60, anchor=north] at (11.0,4.12) {$1$};
\node[slab, black!60, anchor=north] at (16.2,4.12) {$T$};
\node[slab, black!70, anchor=west] at (10.9,3.52) {visible sag: unanchored span drifts};

\draw[arr, rounded corners=3pt] (13.5,3.3) -- (13.5,2.82) -- (6.4,2.82) -- (6.4,2.42);
\node[slab, black!70, anchor=south] at (10.05,2.89)
  {$\anch\leftarrow\anch\cup\{t^{*}\}$;\; $\ifs$-swap $f_{t^{*}}$; regenerate span};

\node[an]  (n0)  at (1.9, 1.9) {};
\foreach \i in {1,...,8} {\node[refr] at (1.9+0.5*\i, 1.9) {};}
\node[nan] at (6.4, 1.9) {};
\foreach \i in {10,...,14} {\node[refr] at (1.9+0.5*\i, 1.9) {};}
\node[an]  (n15) at (9.4, 1.9) {};
\node[lab, anchor=east] at (1.55, 1.9) {$\hat V$, $\anch=\{1,t^{*}\!,T\}$};
\node[slab, black!60, anchor=north] at (5.65,1.5) {regenerated span between $t^{*}$'s neighboring anchors};

\draw[black!60] (10.9,1.0) -- (16.35,1.0);
\draw[black!60] (10.9,1.0) -- (10.9,2.6);
\draw[dashed, black!55] (10.9,1.87) -- (16.35,1.87) node[slab, anchor=west, black!70] {$\tau$};
\draw[genblue!85!black, thick] plot [smooth, tension=0.85] coordinates
  {(11.0,2.43) (11.9,2.28) (12.8,2.12) (13.5,2.38) (14.4,2.08) (15.3,2.22) (16.2,2.45)};
\fill[anchred] (13.5,2.38) circle (0.055);
\node[slab, black!60, anchor=north] at (11.0,0.97) {$1$};
\node[slab, black!60, anchor=north] at (16.2,0.97) {$T$};
\node[slab, okgreen, anchor=north west] at (10.9,0.62)
  {$\min_t s_t \ge \tau \Rightarrow$ \textbf{accept} pair with certificate $(\anch, \min_t s_t)$};
\node[slab, anchred!90!black, anchor=north west] at (10.9,0.18)
  {else if $|\anch| \ge K$ (or no eligible frame) $\Rightarrow$ \textbf{reject} / flag as hard example};

\node[an, minimum width=3mm, minimum height=3mm] at (1.9,0.5) {};
\node[slab, anchor=west] at (2.1,0.5) {IFS anchor};
\node[nan, minimum width=3mm, minimum height=3mm] at (3.8,0.5) {};
\node[slab, anchor=west] at (4.0,0.5) {inserted anchor};
\node[gfr, minimum width=3mm, minimum height=3mm] at (6.2,0.5) {};
\node[slab, anchor=west] at (6.4,0.5) {generated};
\node[refr, minimum width=3mm, minimum height=3mm] at (8.2,0.5) {};
\node[slab, anchor=west] at (8.4,0.5) {regenerated};
\end{tikzpicture}}
\caption{\textbf{The AIA mint loop.} A real video $V_r$ and its pose sequence drive the synthesizer after the boundary frames are IFS-swapped to identity $B$ (top). The generated video is scored per frame against the \emph{real} reference embedding $e_B$; the score curve sags where the unanchored span drifts (right, top). An IFS-swapped anchor is inserted at the sag $t^{*}$ (the guarded $\arg\min$ of Algorithm~\ref{alg:aia}) and the affected span is regenerated (bottom). The loop repeats until the minimum score clears $\tau$ (accept, with a machine-checkable certificate) or the anchor budget $K$ is exhausted (reject, or retain as a flagged hard example).}
\label{fig:pipeline}
\end{figure*}

\section{Adaptive Identity Anchoring}
\label{sec:method}

\subsection{From two anchors to $N$}
\label{sec:nanchor}

In a DiT trained diffusion-forcing style, per-frame noise levels are independent, and conditioning on a frame amounts to clamping its tokens to zero noise throughout sampling~\cite{diffusionforcing}. Nothing in the attention mechanism privileges frames $1$ and $T$; the FLF2V configuration is one point---$\anch=\{1,T\}$---in the space of anchor sets $\anch\subseteq\{1,\dots,T\}$. What \emph{does} privilege the boundary frames is the training distribution: a synthesizer fine-tuned only on first--last conditioning has never seen an interior anchor, and we should expect out-of-distribution artifacts if one is clamped naively at inference.

We therefore propose training (or fine-tuning) the IVS with \emph{randomized anchor sets}: for each training example, sample the anchor count from a distribution concentrated on small values (e.g., $2$--$8$, matching inference) and the positions at random, always retaining the two boundary frames with high probability so that the FLF2V special case remains strongly in-distribution. All anchors are conditioned identically---clamped clean frames---so the modification touches data loading, not architecture; this is why we call the generalization architecturally near-free. The nontrivial risk is not pixel agreement \emph{at} anchors but motion coherence \emph{through} them: keyframe-interpolation work shows that naive two-keyframe conditioning can satisfy the anchors while producing incoherent motion between them~\cite{geninbetween}, and with interior anchors we additionally anticipate an ease-in, pulse artifact in which generated motion decelerates into each anchor and lurches out of it. We hypothesize that randomized anchor training itself mitigates this, because anchors appear at arbitrary motion phases during training rather than exclusively at clip boundaries; as insurance we propose an explicit temporal-coherence check in the mint loop's quality gate (optical-flow smoothness across anchors), so a pair exhibiting anchor pulse fails acceptance like an identity sag.

\subsection{The closed-loop placement algorithm}
\label{sec:loop}

Let $V_r=\{f_1,\dots,f_T\}$ be the real video, $I_B$ the target identity image, $E$ an identity encoder (ArcFace in our instantiation), and $e_B=E(I_B)$ the \emph{reference} embedding---computed from the real photograph of $B$, not from any generated rendition. For a generated video $\hat V=\{\hat f_1,\dots,\hat f_T\}$, the per-frame identity score is
\begin{equation}
s_t \;=\; \cos\!\big(E(\mathrm{crop}(\hat f_t)),\; e_B\big),
\label{eq:score}
\end{equation}
where $\mathrm{crop}(\cdot)$ is the aligned face crop. Let $\anch\subseteq\{1,\dots,T\}$ be the anchor set, $\tau$ the acceptance threshold, $K$ the anchor budget, and $w$ the guard-window radius. Algorithm~\ref{alg:aia} states the loop; Figure~\ref{fig:pipeline} depicts one iteration.

\begin{algorithm}[t]
\small
\caption{Adaptive Identity Anchoring (one mint); \S\ref{sec:texture} adds the spectral channel $h_t$ to the accept test}
\label{alg:aia}
\begin{algorithmic}[1]
\Require $V_r=\{f_1,\dots,f_T\}$; $I_B$; encoder $E$; threshold $\tau$; budget $K$; guard window $w$
\State $e_B \gets E(I_B)$
\State $\anch \gets \{1, T\}$;\quad $a_t \gets \ifs(f_t, I_B)$ for $t\in\anch$
\State $\hat V \gets \ivs\big(\mathrm{pose}(V_r),\, \{a_t : t\in\anch\}\big)$
\Loop
  \State $s_t \gets \cos\big(E(\mathrm{crop}(\hat f_t)),\, e_B\big)$ for $t=1,\dots,T$
  \If{$\min_t s_t \ge \tau$}
     \State \Return \textsc{Accept}$(\hat V, \anch, \min_t s_t)$
  \EndIf
  \If{$|\anch| \ge K$}
     \State \Return \textsc{Reject} \textit{(or flag as hard example)}
  \EndIf
  \State $\mathcal{C} \gets \{\, t : \min_{u\in\anch}|t-u| > w \,\}$
  \If{$\mathcal{C} = \emptyset$}
     \State \Return \textsc{Reject} \textit{(eligible frames exhausted; or flag)}
  \EndIf
  \State $t^{*} \gets \arg\min_{t\in\mathcal{C}}\, s_t$
  \State $\anch \gets \anch\cup\{t^{*}\}$;\quad $a_{t^{*}} \gets \ifs(f_{t^{*}}, I_B)$
  \State \parbox[t]{\dimexpr\linewidth-\algwrapmargin\relax}{\strut regenerate $\hat V$ on the span between $t^{*}$'s neighboring anchors, holding frames outside it fixed \textit{(local-span default; full re-synthesis otherwise)}\strut}
\EndLoop
\end{algorithmic}
\end{algorithm}

\paragraph{Local-span regeneration.} Re-synthesizing the whole clip after each insertion costs one full IVS pass per anchor. Because an inserted anchor alters the constraint problem only between its two neighboring anchors, it suffices to regenerate that span, holding all frames outside it fixed; the span's endpoints are themselves anchors, so the sub-problem has exactly the same form as the whole-clip problem and is in-distribution for a randomized-anchor IVS. With local regeneration, total synthesis cost grows roughly linearly in clip length, not in iterations times clip length. Span-boundary motion coherence is covered by the same temporal check as anchor coherence (\S\ref{sec:nanchor}).

\paragraph{The guard window.} Without the constraint $\min_{u\in\anch}|t-u|>w$, the loop can pile anchors around one stubborn frame: if the IFS anchor at $t^{*}$ is itself weak---IFS is least reliable exactly on profile and occluded views---the sag persists and the next $\arg\min$ lands beside it. The guard window forces the budget to spread; and if the eligible set $\mathcal{C}$ of Algorithm~\ref{alg:aia} empties while the sag persists inside guarded territory, that is evidence the clip is hard for the \emph{anchor generator}, not the synthesizer, and the correct verdict is reject-or-flag rather than further spending. We additionally propose scoring each IFS anchor against $e_B$ \emph{before} insertion and skipping candidates whose anchors fail, so a bad anchor is never installed.

\paragraph{$\tau$ and $K$ as the quality--compute dial.} $\tau$ sets the identity floor a pair must certify; $K$ caps what the factory will spend to reach it. Raising $\tau$ at fixed $K$ trades yield for fidelity; raising $K$ at fixed $\tau$ trades compute for yield. Because every accepted pair carries its certificate $(\anch,\min_t s_t)$, the factory's output is not merely ``pairs'' but pairs with a machine-checkable lower bound on identity fidelity---a property the stock mint does not provide.

\paragraph{The loop as an automatic data filter.} Clips that fail to converge within budget $K$ are exactly the clips on which the synthesizer cannot hold the identity even with help. Discarding them raises mean pair fidelity but narrows coverage; retaining them as \emph{flagged hard examples} preserves coverage while giving the downstream curriculum an explicit difficulty label it currently lacks. We propose flagged retention as the default: the curriculum can schedule certified-easy pairs early and flagged-hard pairs late, and IRL---if still needed---inherits a pre-computed map of where identity is fragile.

\subsection{Drop-in integration with SyncID-Pipe}
\label{sec:integration}

AIA replaces only the mint step. The bidirectional quadruplet structure is unchanged: an accepted $\hat V$ plays the role of $V_g$, both swap directions remain available, the direction regressing onto real $V_r$ pixels retains its role of limiting teacher-manifold inheritance, and background recomposition, the curriculum, and IRL all apply as before. AIA's additional artifacts---per-pair certificates, per-clip difficulty flags, per-frame score curves---strictly extend what the stock pipeline records about its data.

\subsection{Reality-referenced texture restoration}
\label{sec:texture}

RTR treats micro-texture the way AIA treats identity: the generator is never trusted on an axis its objectives cannot see, so the axis is refereed---and here repaired---against real pixels. Three components, all operating on the minted video before acceptance.

\paragraph{Matched re-grain.} Sensor grain is a screen-space field shared by everything the camera captured. We propose estimating each real frame's grain statistics from its own non-face regions (flat patches of background and clothing after edge masking) and applying a matched synthetic grain field to the generated face region of $\hat f_t$. Estimated from the same frame, the field matches the plate's sensor, exposure, and compression state by construction, and per-frame screen-space application makes its motion statistics---static attachment, temporal decorrelation---correct automatically.

\paragraph{Band-split detail transfer.} Face-attached micro-texture exists in the real footage in exactly the right place: on $A$'s skin, under the shot's actual lighting. We propose decomposing the real face crop into a Laplacian-style band stack, warping the highest band onto the generated face along the dense face-attached correspondence, and compositing it over the synthesizer's own high band. The band boundary is set \emph{above} identity-bearing scales: freckles, moles, scars, and wrinkle geometry remain whatever the synthesizer renders for the swapped-in identity $B$; only scale-anonymous micro-texture---the component no identity encoder or human can attribute to a person---transfers from the real footage (\S\ref{sec:limitations} treats the leakage risk when this separation fails).

\paragraph{A second acceptance channel.} Alongside $s_t$ (Eq.~\ref{eq:score}), define for each frame a spectral score on the aligned face crop under a skin mask $M_t$---estimated on the real frame $f_t$ and applied to both crops, so numerator and denominator have matched support---
\begin{equation}
h_t \;=\; \log \frac{E_{>f_c}\big(M_t \odot \mathrm{crop}(\hat f_t)\big)}{E_{>f_c}\big(M_t \odot \mathrm{crop}(f_t)\big)},
\label{eq:spectral}
\end{equation}
where $E_{>f_c}(\cdot)$ is band energy above the cutoff $f_c$, matched to the transfer band boundary. The real frame's own face is the yardstick---same scene, lens, and sensor---so $A$'s skin sets the high-band energy a real face carries in this footage. Acceptance in Algorithm~\ref{alg:aia} then requires both channels, $\min_t s_t \ge \tau$ \emph{and} $\max_t |h_t| \le \epsilon$---two-sided, because too little high-band energy is the beauty filter and too much is synthetic noise. On a spectral-only failure the insertion rule switches channels, $t^{*} \gets \arg\max_{t\in\mathcal{C}} |h_t|$, so the regenerated span covers the spectral sag and receives fresh synthesis, a texture-honest anchor (below), and a fresh RTR pass---the only operations that can change $h_t$ (RTR is deterministic; frames outside the span stay fixed); budget, guard window, and reject-or-flag apply unchanged. The identity referee is the real photograph of $B$; the texture referee is the target footage's own spectrum. Neither is the generator.

\paragraph{Synergy with anchoring.} Anchors are IFS edits of single real frames made at image resolution, never passing through the video VAE: outside the edited face they are real pixels, grain included, and inside it an image-level swap retains far more high-frequency content than video-latent synthesis. We propose applying RTR to each anchor before insertion, making every anchor texture-honest; anchor density then raises the fraction of the clip constrained by real-textured supervision, and the two mechanisms compound: one policy, two axes, one loop.

\section{Why adaptive anchoring should work}
\label{sec:theory}

\paragraph{1. Supervision asymmetry makes pair fidelity the ceiling.}
Pure generation is self-supervised at internet scale; transfer is not. When supervision is manufactured, the student's attainable quality is bounded by the fidelity of the manufactured pairs: a student trained to map $(I_g, V_r)\!\to\!V_g$ can at best learn the identity that $V_g$ actually carries, drift included. DreamID-V's authors implicitly acknowledge this bound twice: by regressing the forward direction onto real pixels, and by adding IRL to repair temporally localized identity failures after the fact~\cite{dreamidv}. Anything that raises minted-pair identity fidelity raises the ceiling for the entire pipeline downstream of the mint, independent of student architecture.

\paragraph{2. Constraint density interpolates between transfer and generation.}
Transfer is dense constraint satisfaction: match reality at every pixel of every frame. Unconditional generation is unconstrained sampling on the model's own manifold. An anchored mint sits between the poles: $k$ anchors force the trajectory to touch reality (as rendered by IFS) $k$ times, and between touches the synthesizer relaxes toward its manifold. Writing $g(t)=\min_{u\in\anch}|t-u|$ for the anchor gap at frame $t$, we hypothesize that expected identity error $\mathbb{E}[1-s_t]$ grows monotonically with $g(t)$---a \emph{drift-versus-gap curve} directly measurable on the existing two-anchor IVS (\S\ref{sec:experiments}, E1). If the curve is increasing, a fixed anchor budget is a coverage problem, and uniform placement is optimal only when drift is translation-invariant along the clip. It plausibly is not: occlusions, profile excursions, and expression peaks are events, not stationary noise. The score curve $s_t$ reveals where the invariance breaks per clip, and adaptive placement spends the budget exactly there.

\paragraph{3. Reality as the referee.}
Within the mint, the synthesizer is graded almost entirely by its own inputs: pose agreement against the extracted skeleton, background and layout against the conditioning video---the generator grades its own homework. Identity appearance between anchors is the sole attribute with no dense reference---the retargeted landmarks densely constrain $B$'s coarse facial geometry but carry no appearance evidence---and it is precisely the attribute on which manifold relaxation is invisible to the generator: drifting toward a more ``generatable'' face \emph{reduces} the synthesizer's internal surprise while destroying the property the pair exists to teach. The loop's score (Eq.~\ref{eq:score}) is computed against $e_B$ from the real photograph of $B$---not against the IFS teacher's rendition and not against any generated frame---so the one unreferenced axis acquires an external referee. This is the same signal DreamID-V already trusts inside IRL~\cite{dreamidv}; AIA moves it upstream, from re-weighting a student's gradients after bad data is minted to preventing the bad mint, where the fix (regenerate a span with one more anchor) is local, cheap, and verifiable. The principle generalizes: the generator may be permitted to grade its own homework on every axis \emph{except} the ones its objectives are blind to. Identity and micro-texture are exactly those axes---identity because manifold relaxation lowers the generator's internal surprise, texture because no loss or reward in the pipeline prices it (\S\ref{sec:beautyfilter})---and both therefore require referees anchored in real pixels: the reference photograph of $B$ for identity, the target footage's own spectrum for texture (Eq.~\ref{eq:spectral}).

\paragraph{4. Targeted coverage of the hard tail.}
DreamID-V's reported failure profile---identity similarity degrading on profile views and intense motion, IRL trained on data selected for high identity variance~\cite{dreamidv}---indicates that minted pairs are least reliable precisely where the task is hardest. Those are also the regions farthest, in constraint terms, from the two boundary anchors: a mid-clip profile excursion is maximally distant from both touches of reality. AIA allocates anchors, and therefore supervision fidelity, preferentially to those regions; each inserted anchor additionally deposits a pixel-exact, IFS-quality supervision point \emph{inside} the pair, where the stock mint provides such points only at the boundaries. We hypothesize that students trained on AIA-minted data need less repair from IRL, shrinking its marginal benefit, because the defect IRL targets is partially removed at the source.

\section{Proposed experiments}
\label{sec:experiments}

\looseness=-1
Hypotheses are stated ex ante. Throughout, evaluation identity encoders are \emph{disjoint} from the loop's scorer (e.g., score with ArcFace~\cite{arcface}; evaluate with CurricularFace~\cite{curricularface} and a third held-out encoder) so that the loop cannot trivially certify itself.

\paragraph{E1: Drift-versus-gap on the stock two-anchor IVS.}
Before anything else, validate the core premise. On held-out real clips, run the existing two-anchor mint, compute $s_t$ for every frame, and regress $1-s_t$ on the anchor gap $g(t)$, stratified by pose (yaw bins), occlusion, and clip length. \emph{Hypothesis:} identity error increases with $g(t)$, with the steepest growth in profile and occlusion strata. \emph{A negative result}---a flat curve---would mean anchor density is not the operative lever and AIA reduces to its filtering role; that would redirect effort toward the conditioning pathway (e.g., identity-bearing dense conditions) rather than anchor placement.

\paragraph{E2: Uniform versus adaptive placement at matched budgets.}
Fine-tune the IVS with randomized anchor sets (\S\ref{sec:nanchor}); then, at each budget $K\in\{2,3,5,8\}$, mint the same clip set with (a)~uniformly spaced anchors and (b)~Algorithm~\ref{alg:aia}, and compare $\min_t s_t$, mean $s_t$, acceptance rate at fixed $\tau$, and flow-smoothness across anchors. \emph{Hypothesis:} adaptive placement dominates uniform at every $K>2$ on minimum identity score, with the largest margin on clips containing localized hard events; uniform and adaptive converge on easy clips. \emph{A negative result} (uniform $\approx$ adaptive) would mean drift is spatially predictable but not clip-specific---still supporting $N>2$ anchors, but removing the case for the closed loop's extra scoring passes.

\paragraph{E3: Student training on AIA-minted versus stock-minted pairs.}
Hold the student architecture, data volume, curriculum, and training schedule fixed; vary only the mint (stock two-anchor versus AIA at a chosen $\tau, K$). Evaluate on an IDBench-V-style protocol~\cite{dreamidv}: identity similarity to the source reference under multiple encoders plus its frame-wise variance, pose and expression error versus the driving video, temporal quality via smoothness metrics and FVD~\cite{fvd,vbench}, strata for profile, occlusion, and long clips, and a held-out generic split as the regression alarm for attribute preservation. \emph{Hypothesis:} identity metrics improve, concentrated in the hard strata; attribute preservation and the generic split stay within noise. \emph{A negative result} with E1--E2 positive would locate the bottleneck in the student (capacity or conditioning), not the data---itself a valuable decomposition, since it would say the pair-fidelity ceiling is not yet binding.

\paragraph{E4: Interaction with Identity-Coherence RL.}
Train students in a $2\times2$ design---\{stock, AIA\} data $\times$ \{with, without\} IRL. \emph{Hypothesis:} IRL's marginal benefit (its with-minus-without delta on identity metrics and on frame-wise variance) shrinks on AIA-minted data. \emph{A negative result}---undiminished IRL benefit---would mean IRL repairs something anchors cannot reach (e.g., student-side sampling variance rather than data-side drift): the mechanisms are complementary and both stay.

\paragraph{E5: Compute accounting.}
A data factory's unit of merit is accepted pairs per GPU-hour at a given certified fidelity. Measure, at each $K$ and $\tau$: IFS calls, IVS passes (full versus local-span regeneration), scoring passes, acceptance rate, and end-to-end throughput; report the fidelity--throughput Pareto front against the stock mint's single point. \emph{Hypothesis:} with local-span regeneration, AIA reaches materially higher certified fidelity at modest throughput cost, and the loop's scoring overhead is negligible next to synthesis. \emph{A negative result}---superlinear cost growth from repeated regeneration on hard clips---would argue for a cheaper variant: one scoring pass, batch insertion of all anchors below $\tau$ in a single second synthesis, no iteration.

\paragraph{E6: Texture restoration ablation.}
Mint the same clip set with and without matched re-grain and band-split transfer (each alone and combined), with the spectral channel logging-only so acceptance does not confound the comparison. Measure the radially averaged high-band energy ratio between generated and real face crops on matched skin masks (the masks $M_t$ of Eq.~\ref{eq:spectral}), LPIPS~\cite{lpips} against the real-frame crops at anchor positions, and a human study in which raters judge which of two matched crops shows the ``beauty-filter'' look, with real--real pairs as controls. \emph{Hypothesis:} restoration moves the energy ratio toward unity and drives beauty-filter identification toward the control error rate, with $s_t$ unchanged. \emph{A negative result}---raters still flag smoothing at matched band energy---would mean energy is too weak a texture summary, and the channel should compare richer statistics (band kurtosis, spatial autocorrelation) instead.

\paragraph{E7: Smoothing across the flywheel.}
Emulate the synthetic-data flywheel: train a generation-1 student on minted pairs, mint new pairs with it, train generation-2 on those, and iterate---once with the texture channel and RTR active, once without. Track the high-band energy ratio of student output across generations. \emph{Hypothesis:} without the texture channel the ratio decays generation over generation (cause~(e) of \S\ref{sec:beautyfilter} compounds); with it the ratio holds. \emph{A negative result}---no decay even without the channel---would mean the flywheel concern is overstated and texture repair is a one-time correction rather than a compounding necessity.

\section{Limitations}
\label{sec:limitations}

\paragraph{The scorer becomes a single point of failure.} The loop optimizes $\min_t s_t$; whatever the identity encoder cannot see, the factory will not fix, and scorer blind spots become student blind spots---Goodhart's law applied to ArcFace. Known encoder weaknesses (extreme yaw, heavy occlusion, low resolution) overlap suspiciously with the very regions AIA targets. Mitigations we propose: ensemble the loop scorer over heterogeneous encoders~\cite{arcface,curricularface}, require agreement for acceptance, and hold out one encoder exclusively for evaluation (\S\ref{sec:experiments}).

\paragraph{Added mint cost.} Each accepted pair may cost up to $K-2$ additional IFS calls, several span regenerations, and $O(K)$ scoring passes. Local-span regeneration bounds the synthesis cost, but at factory scale the overhead is real and is the price of the certificate; E5 quantifies whether it is worth paying.

\paragraph{Anchor pile-up on genuinely hard clips.} The guard window spreads the budget but cannot rescue a clip whose difficulty is intrinsic (continuous occlusion, sustained extreme profile). Such clips exhaust $K$ and exit as rejects or flags; if rejected, the accepted distribution shifts toward easy content, quietly undoing the hard-tail coverage argument of \S\ref{sec:theory}. The flagged-retention default exists precisely to keep the hard tail in the data.

\paragraph{IFS quality bounds anchor quality.} Anchors are IFS outputs; where the IFS model fails, AIA installs a bad constraint and then trusts it. Garbage anchors, garbage pairs. Pre-insertion anchor scoring (\S\ref{sec:loop}) mitigates but does not eliminate this, since the anchor scorer shares blind spots with the loop scorer.

\paragraph{Inherited synthesizer limitations.} AIA improves where the IVS is anchored, not what the IVS can do: resolution ceilings, clip-length limits, motion-class gaps, and through-anchor coherence artifacts~\cite{geninbetween} of the underlying FLF2V model pass through to every minted pair.

\paragraph{Band-split identity leakage.} The detail transfer assumes identity lives below the band boundary and anonymity above it. The assumption is imperfect: some high-frequency content \emph{is} identity---pore-distribution idiosyncrasies, fine wrinkle signatures---and transplanting $A$'s highest band onto $B$'s face risks bleeding $A$ into exactly the signal the identity referee certifies. Sweeping the boundary while measuring the $s_t$ response (E6's protocol accommodates this) is a prerequisite to deployment; when in doubt, the boundary must move up, sacrificing texture for identity.

\paragraph{Goodharting on spectral statistics.} The texture channel prices band energy, so the corresponding failure mode is a generator---or a restoration stage---that adds noise-shaped energy instead of skin. The two-sided threshold rejects gross overshoot, but energy alone cannot distinguish matched-energy synthetic noise from real pores. Richer statistics raise the bar without removing the risk; E6's human study is the backstop, and the held-out-referee discipline of \S\ref{sec:experiments} applies here too: evaluation spectra and raters, never the loop's own statistic, decide.

\section{Ethics statement}
\label{sec:ethics}

Face swapping is dual-use, and a proposal that improves synthetic paired supervision improves the realism of face-swapped video; we do not consider that fact incidental. The texture mechanism in particular would remove one of the commonly used visual tells of synthetic faces---the over-smoothed, beauty-filtered look---which strengthens, rather than weakens, the obligations that follow. Any implementation of AIA must obtain informed consent from all source identities appearing in $V_r$ and all reference identities $I_B$, and must be restricted to footage the operators are authorized to process. Synthetic media produced by the factory or by students trained on it should carry disclosure---visible labeling where appropriate, provenance watermarking always---consistent with DreamID-V's own recommendation to mark generated videos as AI-generated. We note that DreamID-V's released assets are provided for academic research and technical demonstration purposes only, under a click-through license prohibiting malicious, privacy-violating, or misleading applications and requiring explicit consent from identifiable individuals; this proposal targets research use under the same constraints, and any system built on those assets inherits those terms. Finally, AIA's certificates cut both ways constructively: certified synthetic pairs with known anchor sets are also high-quality training and evaluation material for forgery detection, and we encourage sharing such corpora with the detection community under the same consent and licensing constraints.

\section{Conclusion}

SyncID-Pipe demonstrated that video face swapping's missing ground truth can be manufactured by anchoring a pose-driven synthesizer to two IFS-swapped frames. We have argued that the number two is a default, not a law: anchor density is a controllable quality dial, the right controller is a closed loop refereed by the real reference identity, and the loop's verdicts are themselves valuable---as certificates, difficulty labels, and filters. The referee pattern is general: micro-texture, the other axis every objective in the factory is blind to, gets the same treatment, with the footage's own spectrum standing where the reference photograph stands for identity. The proposal is deliberately falsifiable, and E1 is cheap: a drift-versus-gap curve on the existing two-anchor synthesizer would, by itself, tell the community whether the dial exists before anyone spends the compute to turn it.

\end{document}